\documentclass[conference]{IEEEtran}
\IEEEoverridecommandlockouts
\usepackage{cite}
\usepackage{amsmath,amssymb,amsfonts}
\usepackage{algorithmic}
\usepackage{graphicx}
\usepackage{textcomp}
\usepackage{subfigure}
\usepackage{xcolor, algorithm}

\newcommand{\obs}{\text{obs}}
\newcommand{\safe}{\text{safe}}
\newcommand{\HD}{\text{HD}}
\newcommand{\old}{\text{old}}

\def\BibTeX{{\rm B\kern-.05em{\sc i\kern-.025em b}\kern-.08em
    T\kern-.1667em\lower.7ex\hbox{E}\kern-.125emX}}
\begin{document}

\title{Decentralized Adaptive Formation via Consensus-Oriented Multi-Agent Communication\\

\author{\IEEEauthorblockN{Yuming Xiang, Sizhao Li, Rongpeng Li, Zhifeng Zhao and Honggang Zhang}
}
\thanks{Y. Xiang, S. Li, and R. Li are with College of Information Science and Electronic Engineering, Zhejiang University (email: \{xiangym, liszh5, lirongpeng\}@zju.edu.cn).} 
\thanks{Z. Zhao and H. Zhang are with Zhejiang Lab as well as Zhejiang University (email: \{zhaozf, honggangzhang\}@zhejianglab.com).}
}

\maketitle

\begin{abstract}
Adaptive multi-agent formation control, which requires the formation to flexibly adjust along with the quantity variations of agents in a decentralized manner, belongs to one of the most challenging issues in multi-agent systems, especially under communication-limited constraints.
In this paper, we propose a novel Consensus-based Decentralized Adaptive Formation (Cons-DecAF) framework. Specifically, we develop a novel multi-agent reinforcement learning method, Consensus-oriented Multi-Agent Communication (ConsMAC), to enable agents to perceive global information and establish the consensus from local states by effectively aggregating neighbor messages.
Afterwards, we leverage policy distillation to accomplish the adaptive formation adjustment.
Meanwhile, instead of pre-assigning specific positions of agents, we employ a displacement-based formation by Hausdorff distance to significantly improve the formation efficiency.
The experimental results through extensive simulations validate that the proposed method has achieved outstanding performance in terms of both speed and stability.
\end{abstract}


\section{Introduction}
Recently, multi-agent systems (MAS) such as unmanned aerial vehicle (UAV) swarms and Internet of Vehicles flourish due to the flexibility and robustness. Belonging to one of the most fundamental issues in MAS, formation control attracts significant research interest, and multi-agent reinforcement learning (MARL)-based approaches \cite{orr2023multi} emerge with remarkable performance. 
However, contingent on a communication network or leader-follower assumption, these MARL approaches generally face some communication-performance dilemma. For example, \cite{guan2022efficient} unveils that a globally shared observation might generate a significant amount of redundant information, and even yields less competitive result than the case with partial local observation only. Therefore, it becomes critical to design some consensus establishment algorithms to effectively guide the interaction between agents. In that regard, conventional algorithms\cite{amirkhani2022consensus} generally formulate an optimization problem and utilize the control theory to produce a solution. Nevertheless, these algorithms lack the essential flexibility, and can not be easily merged into RL methods.

With the framework of MARL, the centralized training with decentralized execution (CTDE) architecture acts as a foundational solution \cite{yu2022surprising}. For example, many variants of CTDE \cite{das2019tarmac,niu2021multi,wang2021tomc} have been proposed to ameliorate the execution performance by devising a communication module to allow agents to exchange their local information during the training in an explicit or implicit manner. However, 
attributed to a black-box deep neural network, the communication module in \cite{das2019tarmac,niu2021multi} only transmit the local information in a blunt manner, which could not unleash its potential to the full extent (e.g., unable to infer and forward global information during the execution) and fails to filter the meaningful communication content, thus being less competent to handle complex scenarios. 

As a remedy, the opponent modeling approach \cite{wang2021tomc} interprets the communication content as the speculated future actions of other agents. But in partially observable scenarios, this approach suffers from the speculation inconsistency, as individual agents make different speculations according to their own local observations. Thus, it is still challenging to derive some consensus by applying the opponent modeling approach, and perceive consistent information from diversified limited local information. 
On the other hand, the consensus establishment module shall be able to adapt to abrupt environmental changes (e.g., the change in the number of a UAV fleet, some moving obstacles). It implies that respective positions in the formation can not be specifically calibrated (e.g., simply rotatable rings) and assigned to agents \cite{yan2022relative}, since a pre-assigning-based approach inevitably limits the potential applications and incurs severe inefficiency towards an urgently re-organized formation. To sum up, towards adaptive decentralized formation, it remains worthwhile to design a MARL framework to capably reach the consensus in the communication-limited partial observation environment.

In this paper, towards adaptive decentralized formation control in partial observed environment, we propose a Consensus-based Decentralized Adaptive Formation (Cons-DecAF) framework on top of Consensus-oriented Multi-Agent Communication (ConsMAC). In particular, in order to tackle with the global collaboration problem among agents with local observations, we incorporate ConsMAC into Multi-Agent Proximal Policy Optimization (MAPPO) \cite{yu2022surprising}-based CTDE architecture. Furthermore, we take advantage of supervised learning-based policy distillation \cite{yan2022relative,rusu2015policy} to merge diverse learned policies for different agent quantities. Compared with the existing works in the literature, the contribution of our paper can be summarized as follows.

\begin{itemize}
\item We introduce an attention-empowered ConsMAC methodology to unanimously estimate the global information (i.e., the consensus) from different local observations and carefully calibrate exchanged information via a Consensus Establishment (CE) module, which is trained by supervised learning with global information as labels. 
Therefore, the ConsMAC works in a more effectiveness-driven manner, and could significantly enhance the interpretability of the communication module. 
\item We develop a Cons-DecAF framework for distributed formation control in communication-limited environment. In particular, instead of centralized location assignment, we adopt Hausdorff-Distance (HD)\cite{pan2022flexible}-oriented multi-policy-distilled ConsMAC for adaptive formation, which is capable to adopt to agent quantity changes. 

\item We verified the effectiveness and superiority of our framework through extensive simulations in the multi-agent particle environment \cite{lowe2017multi} and the quadcopter-physical-model-based UAV simulation environment \cite{panerati2021learning}.
\end{itemize}

The remainder of the paper is organized as follows. Sec. \ref{s2} presents the system models and formulates the problem. Sec. \ref{s3} elaborates the details of our proposed Cons-DecAF framework. In Sec. \ref{s4}, we introduce the experimental results and discussions. Finally, Sec. \ref{s5} concludes the paper.

\section{System Model and Problem Formulation}\label{s2}
\subsection{System Model}
\begin{figure}[tbp]
\centering
\includegraphics[width=0.35\textwidth]{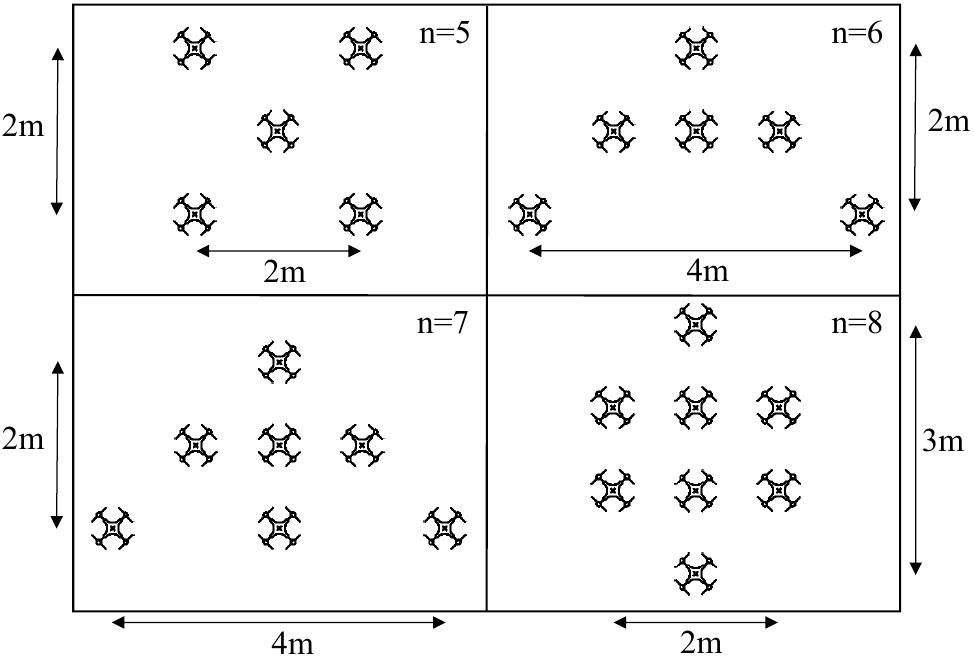}
\vspace{-1.0em}
\caption{An illustration of pre-defined formations corresponding to different number of active UAVs (i.e., $n=5\sim 8$).}
\label{forma}
\vspace{-1.5em}
\end{figure}

We primarily consider a Decentralized Adaptive Formation (DecAF) control problem, wherein a set $\mathcal{N}$ of UAVs shall constitute some pre-defined formation along with moving towards the destination $\Delta_p$ in a communication-limited decentralized manner. Notably, some UAVs might be inactive due to some external reasons (e.g., battery life), and there exist $n$ active UAVs where $n \in [N_{\min}, N_{\max}]$ with $N_{\min}>0$ and  $N_{\max} = \left | \mathcal{N} \right |$. Furthermore, the formation $\Delta_f^n$ is $n$-dependent, as illustrated in Fig. \ref{forma}. For simplicity of representation, the set of active UAVs for a time-step $t$ is denoted as $\mathcal{A}^{(t)}$.
Meanwhile, for each active agent $i \in \mathcal{A}^{(t)}$, the position and velocity are respectively represented by $\mathbf{p}_i^{(t)}=(p_{x_i}^{(t)},p_{y_i}^{(t)})$ and $\mathbf{v}_i^{(t)}=(v_{x_i}^{(t)},v_{y_i}^{(t)})$\footnote{For an inactive UAV, an all-zero vector is used to represent the position and velocity.}.  

In order to accomplish the formation control task, we formulate the problem as a Decentralized Partially Observable Markov Decision Process (Dec-POMDP) \cite{xiao2023stochastic}.
Specifically, 
agent $i$ obtains some direct observation $\mathbf{o}_{i}^{(t)}=\{(\mathbf{p}_j^{(t)},\mathbf{v}_j^{(t)})| \forall j\in \xi ^{(t)}_i\}$ through sensors and receives exchanged messages $\{ \mathbf{m}_j^{(t)}| \forall  j\in \xi ^{(t)}_i\}$\footnote{The detailed procedure to acquire these messages shall be discussed later.}, where $\mathbf{m}_j^{(t)}$ is a learned latent vector to be communicated and $\xi _i^{(t)}$ denotes the set of UAVs $j\in \mathcal{A}^{(t)}$ satisfying that the Euclidean distance $d(\mathbf{p}_i^{(t)},\mathbf{p}_j^{(t)})$ is less than a maximum observation distance $\delta_{\obs}$ (i.e., $d(\mathbf{p}_i^{(t)},\mathbf{p}_j^{(t)}) < \delta_{\obs}$). Therefore, the global state $\mathbf{s}^{(t)}=(\{ (\mathbf{p}_i^{(t)}, \mathbf{v}_i^{(t)})|\forall i\in \mathcal{N} \},\{ \mathbf{m}_j^{(t)}|\forall j\in \mathcal{A}^{(t)} \}) \in \mathcal{S}$ encompasses local states $\textbf{z}_i^{(t)} = (\mathbf{o}_i^{(t)}, \{ \mathbf{m}_j^{(t)}|j\in \xi ^{(t)}_i \} )\in \mathcal{Z}$ of all agents. Notably, by definition, the global state is different from the global observation $\mathbf{o}_{\text{g}}^{(t)}=\{(\mathbf{p}_i^{(t)},\mathbf{v}_i^{(t)})|\forall i\in \mathcal{N}\}$. 
Based on $\textbf{z}_i^{(t)} = (\mathbf{o}_i^{(t)}, \{ \mathbf{m}_j^{(t)}|\forall  j\in \xi ^{(t)}_i \} )$, each agent 
adjusts its acceleration $\mathbf{u}_i^{(t)}=(u_{x_i}^{(t)},u_{y_i}^{(t)}) \in \mathcal{U}$ following individual policy $\pi_i(\cdot |\textbf{z}_i^{(t)}):\mathcal{Z}\times \mathcal{U} \rightarrow [0,1]$, 
which aims 
to reach the formation $\Delta_f^{\hat{n}_i^{(t)}}$ corresponding to an estimated number of active UAVs $\hat{n}_i^{(t)}$. 
Following the joint action $\textbf{U}^{(t)}=\{\mathbf{u}_i^{(t)}|\forall i\in \mathcal{A}^{(t)}\}$, the environment enters into state $\textbf{s}^{(t+1)}$ according to the transition probability 
$\mathcal{P}(\textbf{s}^{(t+1)}|\textbf{s}^{(t)},\textbf{U}^{(t)}):\mathcal{S}\times \mathcal{U} \times \mathcal{S} \rightarrow [0,1]$.
The reward function $r( \textbf{s}^{(t)},\textbf{U}^{(t)}):\mathcal{S}\times{\mathcal{U}}\rightarrow \mathbb{R}$, which shall be discussed in-depth in the next subsection, is shared by all agents. 
Agents need to maximize the discounted accumulated reward $J=\mathbb{E} [{\textstyle \sum_{t}}\gamma ^t r^{(t)}]$, where $\gamma \in [0,1]$ is a discount factor.

\subsection{Reward Design}
In this subsection, we focus on designing the reward function for a fixed number of UAVs during the centralized training, and leave the implementation of a distributed execution in Sec. \ref{s3}. In particular, the developed reward function consists of the following parts.
Beforehand, for simplicity of representation, at each time-step $t$, we use $\bar{\mathbf{p}}^{(t)}$ and $\textbf{P}^{(t)}=(\mathbf{p}_1^{(t)}-\bar{\mathbf{p}}^{(t)},\cdots ,\mathbf{p}_n^{(t)}-\bar{\mathbf{p}}^{(t)})$ to denote the center point and the relative positions of agents.
Besides, we take the reward from the previous step into account to better reflect the formation and navigation trend.

\subsubsection{Formation reward}In order to compute the maximum individual movement distance required for agents to form an ideal topology, the HD \cite{pan2022flexible} is leveraged and a formation reward $r_f^{(t)}$ can be obtained as
\begin{equation}
\label{eq:rform}
    r_f^{(t)}=-d_{\HD}(\textbf{P}^{(t)},\Delta_f^n) - \omega_{1} r_f^{(t-1)},
\end{equation}
where the HD between two formations $\mathcal{T}_1$ and $\mathcal{T}_2$ is defined as $d_{\HD}(\mathcal{T}_1,\mathcal{T}_2)=\max \left \{h(\mathcal{T}_1,\mathcal{T}_2),h(\mathcal{T}_2,\mathcal{T}_1) \right \}$ with $h(\mathcal{T}_1,\mathcal{T}_2) = \max_{\mathbf{x}\in \mathcal{T}_1} \min_{\mathbf{y}\in \mathcal{T}_2}d(\mathbf{x},\mathbf{y})$. Besides, $r_f^{(0)}=0$ and $\omega_{1}$ is the formation lag coefficient.



\subsubsection{Navigation reward}In order to guide the agents towards the target point, the navigation reward, which simply uses the Euclidean distance between the center point of agents and the destination, can be given by
\begin{equation}
    r_{v}^{(t)}=-d(\bar{\mathbf{p}}^{(t)},\Delta_p)- \omega_{2} r_{v}^{(t-1)},
\end{equation}
where $r_{v}^{(0)}=0$ and $\omega_{2}$ is the navigation lag coefficient.
\subsubsection{Collision penalty}The collision penalty counts the total number of collisions at time-step $t$, and can be formulated as,
\begin{equation}
    r_{c}^{(t)}=\sum_{i\ne j} \left ( d(\mathbf{p}_i^{(t)} ,\mathbf{p}_j^{(t)} )<\delta _{\safe} \right ),
\end{equation}
where $\delta_{\safe}$ is the minimum safety distance between agents.

Therefore, the reward function can be summarized by
\begin{equation}
    r^{(t)} = \omega_{f} r_{f}^{(t)}+ \omega_v r_{v}^{(t)} - \omega_c r_{c}^{(t)},
\end{equation}
where $ \omega_{f}$, $ \omega_{v}$, $\omega_{c}$ are the weights of each part.

\subsection{Problem Formulation}\label{sec:mappo}

In order to train a policy that can be executed in a completely distributed manner, we follow the CTDE architecture and leverage MAPPO \cite{yu2022surprising} for multi-agent formation control. Specifcally, MAPPO combines the single-agent PPO and CTDE, and tries to learn the policy  $\pi_{\theta_i}(\cdot |\textbf{z}_i^{(t)})$ ($\forall i$) and value function $V_\phi(\mathbf{s}^{(t)}):\mathcal{S} \rightarrow \mathbb{R} $ parameterized by $\theta_i$ and $\phi$. Consistent with PPO, MAPPO maintains the old-version $\theta_{i,\old}$ and $\phi _{\old}$,
and uses $\theta_{i,\old}$ to interact with the environment and accumulate the samples. Afterwards, MAPPO calculates the generalized advantage estimation (GAE) \cite{schulman2015high} function $\hat{A}^{(t)} = \sum _{l=0}^{T-t-1} (\gamma \lambda)^l \delta^{(t+l)}$ with $ \delta^{(t)}  =  r^{(t)} + \gamma V_{\phi_{\old}}(\mathbf{s}^{(t+1)})-V_{\phi_{\old}}(\mathbf{s}^{(t)})$, where $\lambda$ is a hyperparameter. Furthermore, $\theta_i $ and $\phi $ are periodically updated to maximize
\begin{equation}
\label{eq:ppo}
    \begin{aligned}
        J_{\pi_i}^{(t)}(\theta_i)  = & \min \left( \beta_i \hat{A}^{(t)},\text{clip}\left(\beta_i, 1-\varepsilon, 1+\varepsilon\right) \hat{A}^{(t)}\right),\\
        J_V^{(t)}(\phi) = &-\left (V_\phi (\mathbf{s}^{(t)})- (\hat{A}^{(t)}+ V_{\phi _{\old}}(\mathbf{s}^{(t)}) \right )^2,
    \end{aligned}
\end{equation}
where $\beta_i = \frac{\pi_{\theta_i}\left(\mathbf{u}_i^{(t)}|\textbf{z}_i^{(t)}\right)}{\pi_{\theta_{i,\old}}\left(\mathbf{u}_i ^{(t)}|\textbf{z}_i^{(t)}\right)}$, and $\varepsilon$ is a hyperparameter.
The final optimization objective of MAPPO can be given by
\begin{equation}
\label{mappo}
    J_{\mathrm{MAPPO}}(\theta, \phi)=\mathbb{E}_{i,t} \left[ J_{\pi_i}^{(t)}(\theta_i )+J_V^{(t)}(\phi)+\alpha H(\pi_{\theta_i} (\cdot |\textbf{z}_i^{(t)})) \right],
\end{equation}
where $\alpha$ is coefficient, and $H$ is the entropy function. Recalling that $\textbf{z}_i^{(t)} = (\mathbf{o}_i^{(t)}, \{ \mathbf{m}_j^{(t)}|j\in \xi ^{(t)}_i \} )$ ($\forall i$),
the communicated information (i.e., $\mathbf{m}$) shall impact the final performance. Meanwhile, classical formulation control relies on a centralized reward and cannot be directly applied into distributed UAVs. Therefore we resort to a consensus establishment module ConsMAC to compute $\mathbf{m}$ and propose a distillation-based distributed solution, which are discussed in-depth in Sec. \ref{s3}.

\begin{figure}[tbp]
    \centering
\includegraphics[width = 0.45\textwidth]{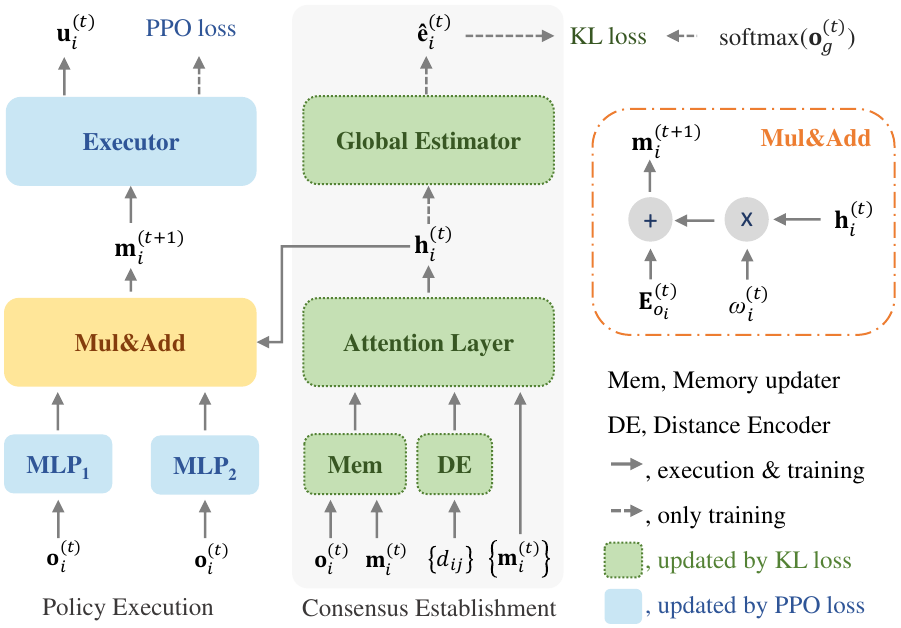}
\vspace{-1.em}
\caption{ConsMAC structure, and the gray part represents the Consensus Establishment (CE) module.}
\label{ConsMAC}
\vspace{-1.5em}
\end{figure} 

\section{The framework of Cons-DecAF}\label{s3}
In this section, we present the details of our Cons-DecAF framework, spanning from the ConsMAC methodology to the distillation-based adaptive distribution formation solution.
\subsection{The ConsMAC Methodology}
Consistent with the conventional MAPPO framework \cite{yu2022surprising,yan2022relative}, the value function $V_\phi$ is implemented by using a MultiLayer Perceptron (MLP) as well. However, significantly different from MAPPO, ConsMAC introduces a novel Consensus Establishment (CE) module and incorporates the established consensus by significantly modifying the policy module, which has been illustrated in Fig.~\ref{ConsMAC}. For convenience, we denote the network realized by MLP as $\mathcal{F}(\cdot)$ in this subsection.

\subsubsection{Consensus establishment module}\
Targeted at effectively aggregating observations $\mathbf{o}$ and messages $\mathbf{m}$ received from neighbors, the CE module is carefully calibrated and encompasses the following parts. Notably, for simplicity of representation, for any UAV (i.e., agent) $i\in \mathcal{A}^{(t)}$, $i_j$ denotes the $j$-th nearest neighbor while $i_0$ indicates itself.

\noindent\textbf{Distance Encoder (DE)}: In resemblance to the word encoding in Transformer\cite{vaswani2017attention}, we use a distance encoder $\mathcal{F}_{\psi_{D}}(d(\mathbf{p}_i^{(t)},\mathbf{p}_{i_j}^{(t)}))$  parameterized by $\psi_D$ to distinguish messages from agents with different distances and correspondingly assign different attention weights.

\noindent\textbf{Memory updater (Mem)}: Each agent $i$ concatenates its own message $\mathbf{m}_{i_0}^{(t)}$ based on the local observation $\mathbf{o}_i^{(t)}$ by a memory updater $\mathcal{F}_{\psi_M}$ parameterized by $\psi_M$, that is,
\begin{equation}
\label{eq:lse3}
    \mathbf{m}'_{i_0}= \mathcal{F}_{\psi_M}([\mathbf{m}_{i_0}^{(t)}||\mathbf{o}_i^{(t)}]),
\end{equation}
where $ \|$ represents the concatenation operation and the dimensions of both $\mathbf{m}_{i_0}^{(t)}$ and $\mathbf{m}'_{i_0}$ are the same. The Mem module acts like a GRU-based module and allows $\mathbf{m}'_{i_0}$ to incorporate historical information. 

\noindent\textbf{Attention Layer}: 
Based on DE and Mem, the input of attention layer can be written by
\begin{equation}
\label{eq:DE}
    \begin{aligned}
        &\mathbf{E}_{m_i}^{(t)} = \left[\mathbf{m}'_{i_0}||\Phi_{d_0}^{(t)}, 
             \dots, \mathbf{m}_{i_k}^{t}||\Phi_{d_k}^{(t)}\right]^{\top},\\
    \end{aligned}
\end{equation}
where $\Phi_{d_j}^{(t)}=\mathcal{F}_{\psi_{D}} (d(\mathbf{p}_i^{(t)},\mathbf{p}_{i_j}^{(t)}))$ and $k=|\xi ^{(t)}_i|$ is the number of neighbors. 
Then by plugging \eqref{eq:DE} into a multi-head attention, each agent can aggregate a latent vector $\mathbf{h}_i^{(t)}$ as
\begin{equation}
\label{eq:AL}
    \begin{aligned}
        \mathbf{h}_i^{(t)} = \operatorname{MultiHead}_{\psi_{A}}
        (\mathbf{E}_{m_i}^{(t)},\mathbf{E}_{m_i}^{(t)},\mathbf{E}_{m_i}^{(t)}),
    \end{aligned}
\end{equation}
where an $h$-head attention is defined as
\begin{equation}
\label{eq:multihead}
    \begin{aligned}
        &\operatorname{MultiHead}( \mathbf{Q},\mathbf{K},\mathbf{V}) = 
        [\mathbf{head}_0 || \dots || \mathbf{head}_h ]\mathbf{W}_{O}.
    \end{aligned}
\end{equation}
Notably $\mathbf{head}_i=\operatorname{Attention}( \mathbf{Q} \mathbf{W}_i^Q,
\mathbf{K} \mathbf{W}_i^K, \mathbf{V} \mathbf{W}_i^V)$ with $\operatorname{Attention}( \mathbf{Q},\mathbf{K},\mathbf{V}) = \operatorname{softmax} 
        \left(\frac{\mathbf{Q}\mathbf{K}^{\boldsymbol{\top}}}{\sqrt{\varepsilon }}\right) \mathbf{V}$, and $\varepsilon $ is a constant. $\mathbf{W}_i^{Q}$, $\mathbf{W}_i^{K}$, $\mathbf{W}_i^{V}$ are the projection matrices for head $i$, and $\mathbf{W}_{O}$ is a parameter matrix as well. 
The attention layer parameterized by $\psi_A$ is used to aggregate the information of neighbors and itself.

\noindent \textbf{Global Estimator}: Different from the work\cite{das2019tarmac,niu2021multi}, we specially encode the global information by a global estimator $\mathcal{F}_{\psi_E}$ parameterized by $\psi_E$, and the estimated state embedding can be written as
\begin{equation}
\label{eq:GE}
    \begin{aligned}
        &\mathbf{\hat{e} } _i^{(t)}  = \mathcal{F}_{\psi_E}({\mathbf{h}_i^{(t)}}).
    \end{aligned}
\end{equation}

We take the true global observation $\mathbf{o}_{g}^{(t)}=\{(\mathbf{p}_j^{(t)},\mathbf{v}_j^{(t)})|\forall j\in \mathcal{N}\}$ as the label for supervised learning, and adopt pointwise Kullback–Leibler (KL) divergence as the loss function. Therefore, the loss function of CE can be formulated as
\begin{equation}
\label{eq:celoss}
    \begin{aligned}
        \mathcal{L}_{\mathrm{CE}}(\Psi)=  \mathbb{E}_{i,t}\left [\sum _x \textbf{e}_{g}^{(t)}(x)\ln \frac{\textbf{e}_{g}^{(t)}(x)}{\mathbf{\hat{e} } _i^{(t)}(x)} \right],
    \end{aligned}
\end{equation}
where $\textbf{e}_{g}^{(t)} = \mathrm{softmax}(\mathbf{o}_{g}^{(t)})$ and $\Psi=[\psi_D,\psi_M,\psi_A,\psi_E]$. 
In other words, the intermediate output of the CE module can implicitly embed the global information. 
Due to the uniqueness of the global information, it can be regarded as the established consensus between agents.

\subsubsection{Policy execution module}\
The Policy Execution (PE) module aims to yield a suitable action corresponding to the observations and established consensus (i.e., $\mathbf{o}_i^{(t)}$ and $\mathbf{h}_i^{(t)}$). In particular, an embedding vector $\mathbf{E}_{o_i}^{(t)}$ and the communication information weight $w_i^{(t)}$ can be obtained by two MLP-based encoders $ \mathcal{F}_{\theta_O}$ and $\mathcal{F}_{\theta_W}$ as
\begin{equation}
\label{eq:lse12}
    \begin{aligned}
        \mathbf{E}_{o_i}^{(t)} = \mathcal{F}_{\theta_O}(\mathbf{o}_i^{(t)}),\quad
        w_i^{(t)} = \mathcal{F}_{\theta_W}(\mathbf{o}_i^{(t)}),
    \end{aligned}
\end{equation}
Afterwards, the message $\mathbf{m}_i^{(t+1)}$ for time-step $t+1$ will be calculated as
\begin{equation}
\label{eq:muladd}
    \begin{aligned}
        \mathbf{m}_i^{(t+1)}=\mathbf{E}_{o_i}^{(t)} + w_i^{(t)} \mathbf{h}_i^{(t)}.
    \end{aligned}
\end{equation}
It can be observed from \eqref{eq:muladd} that the weight $w_i^{(t)}$ will affect the importance of $\mathbf{h}_i^{(t)}$ based on the observation $\mathbf{o}_i^{(t)}$. Furthermore, an Executor $\mathcal{F}_{\theta_E}$ is used to sample an action as the final output by computing the mean of the Gaussian distribution as
\begin{equation}
\label{eq:PE}
    \begin{aligned}
        &\boldsymbol{\mu}_i^{(t)} = \mathcal{F}_{\theta_E}(\mathbf{m}_i^{(t+1)}),\quad
        \mathbf{u}_i^{(t)} \sim \operatorname{Normal}(\boldsymbol{\mu}_i^{(t)}, \sigma),\\
    \end{aligned}
\end{equation}
where $\sigma$ is the variance constant that provides randomness to the agent while exploring the environment, and it gradually decreases during the training. 
As mentioned in Sec. \ref{sec:mappo}, consistent with MAPPO, we can treat
$[\Theta,\Psi]$ as the parameters of the final policy $\pi$ in \eqref{eq:ppo},
where $\Theta=[\theta_O,\theta_W,\theta_E]$.

\subsubsection{Training techniques}\

In a nutshell, the loss function of ConsMAC for a specific number $n$ can be summarized as
\begin{equation}
\label{eq:ConsMACloss}
    \mathcal{L}_{\mathrm{ConsMAC}}(\Theta, \phi, \Psi)=-J_{\mathrm{MAPPO}}(\Theta, \phi) + \mathcal{L}_{\text{CE}}(\Psi),
\end{equation}
In addition, even though $\mathbf{h}_i^{(t)}$ is used as the input of the PE module, the PPO loss doesn't backpropagate to the CE module.
In other words, the CE module is designed as an independent information processing module to provide the global consensus. 
The parameters of our agents are shared to improve the efficiency of learning.

\begin{figure}[tbp]
    \centering
\includegraphics[width = 0.4\textwidth]{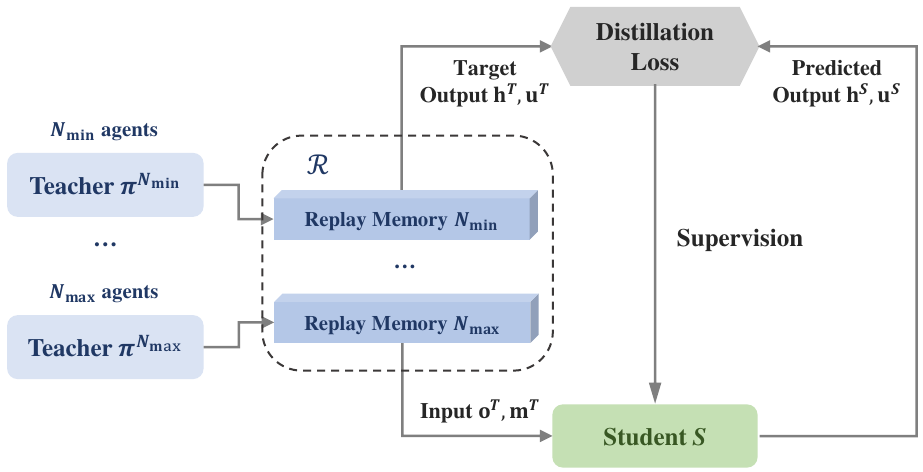}
\vspace{-.8em}
\caption{Adaptive Formation via policy distillation. }
\label{pd}
\vspace{-1.5em}
\end{figure} 

\subsection{Adaptive Formation Control}
Since we focus more on the formation models under different $n\in [N_{\min}, N_{\max}]$ in this subsection, we omit the subscript $i$ for brevity.
After learning the policy $\pi^n$ parameterized by $\Theta^n,\Psi^n$ for a specific $n$ separately, it becomes possible to integrate multiple formation models corresponding to different $n$. In that regard, we take the technique of policy distillation \cite{yan2022relative,rusu2015policy}.
Specifically, as shown in Fig. \ref{pd}, we collect the input and output at multiple time steps of different policy models $\pi^n$ as teacher models by constituting a replay memory $\mathcal{R}= \{\langle\mathbf{z}, \mathbf{u}, \mathbf{h} \rangle^{\pi^{N_{\min}}}
        ,\dots,\langle\mathbf{z}, \mathbf{u}, \mathbf{h} \rangle^{\pi^{N_{\max}}}\}_{\Lambda }$,
where $\Lambda $ is the capacity of replay memory.
Notably, in addition to the regular local states $\mathbf{z}$ and actions $\mathbf{u}$, we store the output of teacher's attention layer $\mathbf{h}$ as well, since $\mathbf{h}$ contributes to retaining the benefits of ConsMAC aggregated information. 
Besides, we align vectors of different formation memories by zero padding if $n < N_{\max}$ and thus the agent number is implicitly indicated in the replay memory. 
In each training episode, memories (i.e., $\mathbf{z}$) from different teacher models are fed into the student model $\pi_S$ 
simultaneously to calculate the corresponding $\mathbf{u}_S$ and $\mathbf{h}_S$. Afterwards, we train $\pi_S$ by minimizing Mean-Squared-Error (MSE) loss
\begin{equation}
\label{eq:pdloss}
    \mathcal{L}_{\text{PD}}(\Theta_S,\Psi_S) = \sum\nolimits_{\langle\mathbf{u},\mathbf{h}\rangle\in \mathcal{R}}(\left \| \mathbf{u} -\mathbf{u}_{S} \right \| ^2_2 + \left \| \mathbf{h}-\mathbf{h}_{S} \right \| ^2_2),
\end{equation}
where $\Theta_S,\Psi_S$ are the parameters of $\pi_S$.
Typically, we use supervised learning to train the student model, which can eventually make similar decision as teacher models.

In short, we summarize the training procedure of Cons-DecAF in Algorithm \ref{Algorithm1}.
\begin{algorithm}[tbp]
\caption{The Training of Cons-DecADF}\label{Algorithm1}
\begin{algorithmic}[1]
    \STATE Initialize the number of agents $N_{\min}$, $N_{\max}$, the length of episodes $L$, variance constant $\sigma$, the student network $\pi^S$ with random parameters $\Theta_S,\Psi_S$ and the replay memory $\mathcal{R}\leftarrow\varnothing$;
    \FOR{$n = \{ N_{\min},\cdots, N_{\max}\}$}
        \STATE Initialize the teacher policy $\pi^n$ and value function with random parameters $\Theta^n,\Psi^n$ and $\phi^n$ respectively;
        \FOR{each train episode}
            \STATE Clone $\Theta_{\old}^{n} \leftarrow \Theta^{n}$, $\Psi_{\old}^{n} \leftarrow \Psi^{n}$, $\phi_{\old}^{n} \leftarrow \phi^{n}$;
            \STATE Initialize the environment corresponding to $n$;
            \FOR{$t= \{ 1,\cdots, L\}$}
                \STATE \label{start} Each agent $i$ obtains a local state $\mathbf{z}_i^{(t)}= (\mathbf{o}_i^{(t)}, \{ \mathbf{m}_j^{(t)}|\forall  j\in \xi ^{(t)}_i \} )$ from the environment;
                \STATE Each agent calculates $\mathbf{h}_i^{(t)}$ by \eqref{eq:AL} and generates the message $\mathbf{m}_i^{(t+1)}$
                by \eqref{eq:muladd} to sample an action $\mathbf{u}_i^{(t)}\sim \operatorname{Normal}(\boldsymbol{\mu}_{i,\old}^{(t)}, \sigma)$ by \eqref{eq:PE};
                \STATE \label{end} Obtain the reward $r^{(t)}$, state value $V_{\phi_{\old}}(\mathbf{s}^{(t)})$ and $\mathbf{s}^{(t+1)}$ at the end of $t$;
            \ENDFOR
            \STATE For each time-step $t$, each agent calculates $\boldsymbol{\mu}^{(t)}_i$, $\mathbf{\hat{e}}^{(t)}_i$ based on $\mathbf{z}_i^{(t)}$ by \eqref{eq:lse3}-\eqref{eq:PE}, and obtains $V_{\phi}(\mathbf{s}^{(t)})$;
            %
            \STATE Update $\Theta^{n}$, $\Psi^{n}$ and $\phi^n$ according to \eqref{eq:ConsMACloss} via Adam optimizer;
        \ENDFOR   
    \ENDFOR
    \STATE Collect tuples $\langle \mathbf{z}, \mathbf{u}, \mathbf{h} \rangle$ by $\pi^{N_{\min}},\cdots,\pi^{N_{\max}}$ respectively and store them in $\mathcal{R}$;
    \FOR{each policy distillation episode}
        \STATE Sample a batch of $\langle \mathbf{z}, \mathbf{u}, \mathbf{h} \rangle$ from $\mathcal{R}$;
        \STATE Calculate $\mathbf{h}_S,\mathbf{u}_S$ based on $\mathbf{z}$ by \eqref{eq:AL} and \eqref{eq:PE};
        \STATE Update $\Theta_S$, $\Psi_S$ according to \eqref{eq:pdloss} via Adam optimizer;
    \ENDFOR
\end{algorithmic}
\end{algorithm}
 
\section{Experimental Results and Discussions}\label{s4}
In this section, we evaluate our method in both multi-agent particle environment (MPE) \cite{lowe2017multi} and gym-pybullet-drones \cite{panerati2021learning}, a simulation environment for multiple quadcopters based on the Bullet physics engine.
The experiments in Sec. \ref{exp:hd} and Sec. \ref{exp:ConsMAC} are first conducted in MPE to demonstrate the effectiveness and superiority of HD-based formation and ConsMAC method over some baselines.
Afterwards, in Sec. \ref{exp:adf}, we validate the Cons-DecAF in gym-pybullet-drones. In particular, corresponding to different number of UAVs, we design different formation topologies as shown in Fig. \ref{forma}. 
Moreover, the Distance Encoder and the Executor consist of a linear layer, while the value function is implemented by a $3$-layer MLP and the rest of MLP-based modules have $2$ layers. Besides, we implement a $4$-head Attention Layer. Moreover, the dimension of $\mathbf{m}_i^{(t)}$ and hidden size of MLPs are $256$ for all $i\in \mathcal{N}$. 
During the experiments, we only consider the formations on the same plane, which implies 
$\mathbf{u}_i^{(t)}$ is $2$-dimensional. 
We utilize some common training techniques in MAPPO such as orthogonal initialization, gradient clipping and value normalization. The model is updated for $350$ episodes, each of which has $100$ time-steps and the coordinates of the UAVs are randomly initialized. 
Adam optimizer is used with a learning rate of $1\times 10^{-4}$.
Finally, we summarize the remaining key parameter settings in Table \ref{tab:para}.

\begin{table}[]
    \centering
    \vspace{-0.5em}
    \caption{The remaining key parameter settings.}
    \label{tab:para}
    \begin{tabular}{|l|l|}
    \hline
     \multicolumn{1}{|c|}{\textbf{Parameters}} & \multicolumn{1}{c|}{\textbf{Value}} \\ \hline
     \multicolumn{1}{|c|}{The Range of $\mathbf{u}$}    & \multicolumn{1}{|c|}{$[-0.5,0.5]$ m/s\textsuperscript{2}} \\ \hline
     \multicolumn{1}{|c|}{The Range of $p_x^0,p_y^0$}      & \multicolumn{1}{|c|}{$[-2,2]$ m}    \\ \hline
     \multicolumn{1}{|c|}{Destination $\Delta_p$}      & \multicolumn{1}{|c|}{$(0,10)$ m}    \\ \hline
    \multicolumn{1}{|c|}{The Range of $\sigma$}      & \multicolumn{1}{|c|}{$[0.01, 0.5]$}     \\ \hline
    \multicolumn{1}{|c|}{Discount Factor $\gamma$}    & \multicolumn{1}{|c|}{$0.8$} \\ \hline
    \multicolumn{1}{|c|}{GAE $\lambda$}      & \multicolumn{1}{|c|}{$0.95$}     \\ \hline
    \multicolumn{1}{|c|}{$(\omega_1,\omega_2,\omega_f,\omega_v,\omega_c)$}      & \multicolumn{1}{|c|}{$(0.9,0.7,10,5,6)$}     \\ \hline
    \end{tabular}
    \vspace{-1.5em}
\end{table}

\subsection{Formation Metric Performance Comparison}\label{exp:hd}
To demonstrate the flexibility and efficiency of the HD metric, we compare it with the classical Point-To-Point (PTP) metric,
in which each target position is specially assigned to one of the UAVs and the formation reward is directly calculated in terms of Euclidean distance.
We evaluate both metrics under the conventional MAPPO settings with $\delta_{\obs}$ set to $3$ m and keeping others the same as HD formation.
By counting the average moving distance and time of agents during the formation process, Table \ref{tab:ptphd} shows the formation cost of $50$ rounds of tests. It can be observed that compared with PTP, HD greatly reduces the overhead of formation (i.e., faster formation and shorter moving distance), as it allows agent to choose its own target point more flexibly instead of simply obeying the assignment.

\begin{table}[]

\centering
\caption{Average moving distance and time steps of point-to-point and hausdorff distance formation}
\label{tab:ptphd}
\begin{tabular}{l|llll}
\hline
No. of agents & $5$-agent & $6$-agent & $7$-agent & $8$-agent \\ 
\hline
PTP-distance  & $1.81$    & $2.47$    & $2.72$    & $2.84$    \\
HD-distance   & $0.76$    & $1.13$    & $1.64$    & $1.37$    \\ 
\hline
PTP-time     & $27.10$   & $41.52$   & $43.00$   & $53.44$   \\
HD-time      & $12.62$   & $17.26$   & $29.02$   & $35.72$   \\
\hline
\end{tabular}
\vspace{-.5em}
\end{table}


\begin{figure}[tbp]
    \centering
\subfigure[$\delta_{\obs}=3\mathrm{m}$]{
\includegraphics[width = 0.38\textwidth]{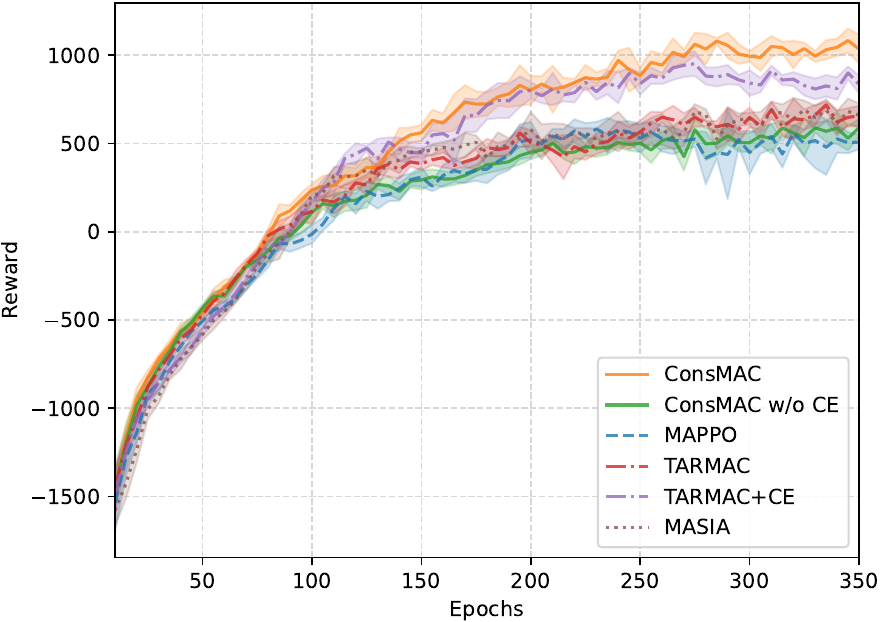}\label{cl3}}
\subfigure[$\delta_{\obs}=2\mathrm{m}$]{
\includegraphics[width = 0.38\textwidth]{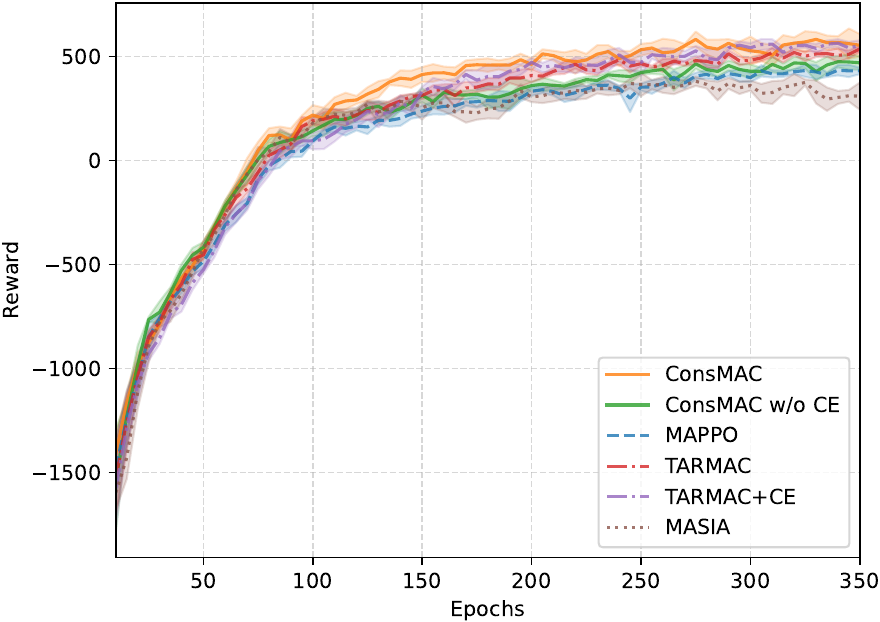}\label{cl2}}
\vspace{-.5em}
\caption{The learning curve of different methods under different $\delta_{\obs}$.}
\vspace{-1.em}
\end{figure} 

\subsection{ConsMAC Performance Experiments}\label{exp:ConsMAC}
In this part, we focus on the impact of different message aggregation methods on the formation model in our comparison.
We take the formation of 7-agent as the example to compare two representative MARL communication methods (i.e., the attention-based message aggregation method TarMAC \cite{das2019tarmac} and the state-of-the-art supervised learning-based information extraction method MASIA \cite{guan2022efficient}) with our ConsMAC module. 
In addition, we add three controlled experiments to reflect the effect of the communication module and our CE module: 
\begin{itemize}
    \item \textbf{MAPPO}: Original MAPPO without communication. 
    \item \textbf{ConsMAC w/o CE}: ConsMAC without the supervised learning loss. 
    \item \textbf{TarMAC+CE}: TarMAC with Global Estimator and supervised learning loss.
\end{itemize}
Fig. \ref{cl3} compares the performance of methods when $\delta_{\obs}=3$ m.
As shown in Fig. \ref{cl3}, the default MAPPO without any communication module performs worst, which validates the effectiveness to add a communication module.
Meanwhile, ConsMAC significantly outperforms other baselines and the introduction of the CE module to ConsMAC greatly contributes to the overall performance improvement.
Besides, it is worth noting that adding a CE module to the ordinary TarMAC apparently boosts the performance as well, which demonstrates the robustness and generality of the CE module. Fig. \ref{cl2} further shows ConsMAC maintains competitive performance in contrast to the declined performance of other methods under more stringent communication limitations ($\delta_{\obs}=2$ m).


\begin{figure}[tbp]
    \vspace{-0.0em}
    \centering
\includegraphics[width = 0.38\textwidth]{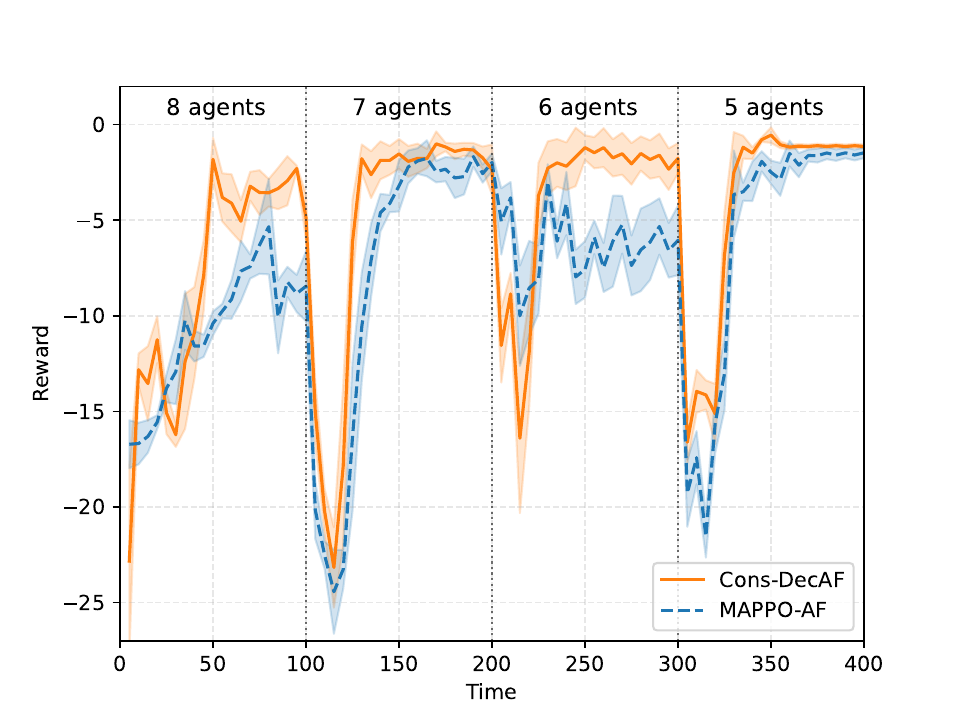}
\vspace{-1.em}
\caption{Curve of formation reward over time, where every 100 time steps a random agent stragglers.}
\label{kd}
\vspace{-1.5em}
\end{figure} 

\subsection{Adaptive Formation Results}\label{exp:adf}
For the adaptive formation, we pre-train four formation models with the number of agents $n$ ranging from $5$ to $8$, and store the corresponding tuples $\langle\mathbf{z}, \mathbf{u}, \mathbf{h} \rangle$ for $60,000$ time-steps as the replay memory $\mathcal{R}$. Besides, Adam optimizer is used with a learning rate of $2\times 10^{-4}$.
Meanwhile, the distilled student model is trained for $200,000$ episodes and the batch size of each episode of sampling is $500$.
We evaluate the MAPPO-based adaptive formation \cite{yan2022relative}, denoted as MAPPO-AF, and our Cons-DecAF in the gym-pybullet-drones environment with $\delta_{\obs}=2$ m. In particular, given the initial existence of $8$ agents, some UAVs are randomly selected and assumed to be no longer observed in turn. In adaptive formation task, agents are supposed to perceive the change of the fleet number itself and adapt the formation policy swiftly.
In order to unify the affects of different formation topologies, we set the maximum value of each formation reward to $0$.
Fig. \ref{kd} illustrates the curve of formation reward over time. It can be observed that the Cons-DecAF can not only quickly switch formations according to the current number of agents, but also achieve more stable performance and superior efficiency than conventional MAPPO, which validates the superiority of our proposed method.

\section{Conclusion}\label{s5}
In this work, we have proposed and validated a novel framework named Consensus-based Decentralized Adaptive Formation (Cons-DecAF).
Specifically, by utilizing supervised learning method, we have designed Consensus-oriented Multi-Agent Communication (ConsMAC) methodology, in which the agent can infer the global information from the local state to implicitly establish the consensus.
Meanwhile, HD-based formation without pre-assigned positions and policy distillation are adopted to achieve a more flexible adaptive formation.
We have conducted extensive experiments to successfully demonstrate the effectiveness \& robustness of our proposed ConsMAC method in communication-limited environment, and verified the superiority of Cons-DecAF in MPE and gym-pybullet-drones.
In the future work, we will carry out intense study in larger-scale formation with higher communication restrictions, and try to implement the approach in a more practical  hardware platform.

\bibliographystyle{IEEEtran}
\bibliography{reference}
\end{document}